\documentclass[10pt, a4paper, twocolumn, showabstract]{naverlabseurope}

\usepackage{booktabs}       % professional-quality tables
\usepackage{amsfonts}       % blackboard math symbols
\usepackage{nicefrac}       % compact symbols for 1/2, etc.
\usepackage{pifont}
\usepackage{xcolor}         % colors
\usepackage{multirow}
\usepackage{microtype}      % microtypography
\usepackage{graphicx} 
\usepackage{amsmath}
\usepackage{amssymb}
\usepackage{mathtools}

\newlength{\smallimage}
        \definecolor{rel}{rgb}{.1,.6,.2}
        \definecolor{nrl}{rgb}{1,1,1}
        \definecolor{qim}{rgb}{1,1,1}

\definecolor{lightgray}{gray}{0.93}

\def\be{\begin{equation}}
\def\ee{\end{equation}}
\def\bea{\begin{eqnarray}}
\def\eea{\end{eqnarray}}
\def\ben{\begin{eqnarray*}}
\def\een{\end{eqnarray*}}

\def\bi{\begin{itemize}}
\def\ei{\end{itemize}}

\newcommand{\btab}[1]{\begin{tabular}{#1}}
\newcommand{\etab}{\end{tabular}}
\newcommand{\ba}[1]{\begin{array}{#1}}
\newcommand{\ea}{\end{array}}

\def\<{\langle}
\def\>{\rangle}



\definecolor{DarkGreen}{rgb}{0.5, 0.9, 0.5}

\def\etal{\emph{et al}.}

\def\name{{\it InConDiff}}

\newcommand{\PAR}[1]{\noindent{\bf{#1.}}}

\newcommand{\hide}[1]{}

\definecolor{Gray}{gray}{0.85}
\definecolor{GrayBorder}{gray}{0.65}
\definecolor{green_cylinder}{rgb}{0.0,0.40,0.0}
\definecolor{blue_cylinder}{rgb}{0.02,0.05,0.75}
\definecolor{way_point}{rgb}{0.56,0.0,1.0}
\newcolumntype{a}{>{\columncolor{GrayBorder}}c}
\title{3D-Consistent Image Inpainting with Diffusion Models}
\titlerunning{3D-Consistent Image Inpainting with Diffusion Models}

\correspondingauthor{[boris.chidlovskii,leonid.antsfeld]@naverlabs.com}

\authors{Leonid Antsfeld, \authsep Boris Chidlovskii}
\affiliations{NAVER LABS Europe}
\contributions{} %$^{\star}$equal contribution}
\website{https://github.com/naver/croco-diffusion}
\websiteref{\href{https://github.com/naver/croco-diff}}

\begin{abstract}
We address the problem of 3D inconsistency of image inpainting based on diffusion models. We propose a generative model using image pairs that belong to the same scene. To achieve the 3D-consistent and semantically coherent inpainting, we modify the generative diffusion model by incorporating an alternative point of view of the scene into the denoising process. This creates an inductive bias that allows to recover 3D priors while training to denoise in 2D, without explicit 3D supervision. Training unconditional diffusion models with additional images as in-context guidance allows to harmonize the masked and non-masked regions while repainting and ensures the 3D consistency. We evaluate our method on one synthetic and three real-world datasets and show that it generates semantically coherent and 3D-consistent inpaintings and outperforms the state-of-art methods.
\end{abstract}

\begin{document}
\maketitle

%%%%%%%%%%%%%%%%%%%%%%%%%%%%%%%%%%%%%%%%%%%%%%%%%%%%%%%%%
%%%%%%%%%%%%%%%%%%%%%%%%%%%%%%%%%%%%%%%%%%%%%%%%%%%%%%%%%
\section{Introduction}
\label{sec:introduction}
\noindent

Image inpainting refers to the task of filling in missing regions within an image~\cite{barnes2009patchmatch,pathak2016context} %, based on non-masked regions
specified by a binary mask~\cite{huang2024diffusion_editing_survey}.
Such inpainted regions need to be semantically consistent %~\todo{reasonable? smooth} 
and harmonized with the rest of the image.
Moreover, inpainting methods need to handle various forms of masks where, in extreme cases, a vast majority of the image is missing. 
Any approach trained with a certain mask distribution can lead to poor generalization to novel mask types~\cite{liu2018image}.

Recent advances in diffusion models led to image inpainting using 
Denoising Diffusion Probabilistic Models (DDPMs)~\cite{ho20_denoising_diffusion,diffusionThermo}. 
The DDPM is trained to denoise the image by reversing a diffusion process. 
Starting from randomly sampled noise, it is iteratively applied for a certain number of steps and produces the final image sample; DDPMs have shown a strong capacity to generate diverse and high-quality 2D images~\cite{ho20_denoising_diffusion,improvedddpm,beatGan}.
Moreover, the problem of %no mask-specific training
generalization to novel masks can be circumvented by inpainting that leverages {\it unconditionally trained} DDPMs~\cite{lugmayr22repaint,Corneanu2024latentpaint}.
Instead of learning a mask-conditional generative model, they condition the generation process by sampling from masked pixels during the reverse diffusion iterations. The unconditional DDPM is used as a generative prior to harmonize information between masked and non-masked regions, by performing forward diffusion of non-masked regions and reverse diffusion of masked ones, operating in pixel~\cite{lugmayr22repaint} or latent spaces~\cite{Corneanu2024latentpaint}.
This allows the network to generalize to any mask during inference.

%-------------------------------------------------------
%----------------------------------------------
\begin{figure}[t!] \centering  
    \begin{tabular}{cccc}
        a)Original & b)Masked   & c)In-context  & d)\name \\
        image       & occlusions  &   image       & inpainting  \\
    \end{tabular}
    \includegraphics[width=0.99\columnwidth]{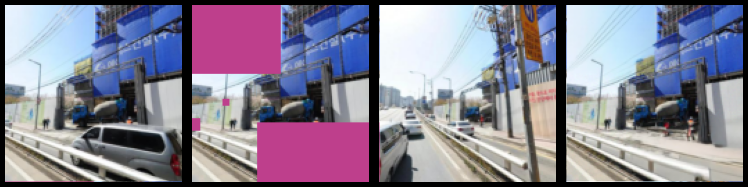}\\
    \includegraphics[width=0.99\columnwidth]{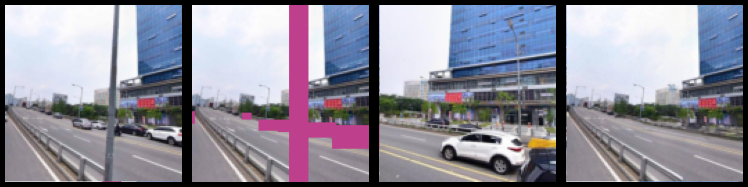}\\
    \includegraphics[width=0.99\columnwidth]{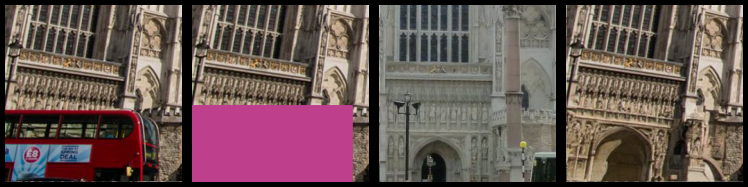}\\
    \includegraphics[width=0.99\columnwidth]{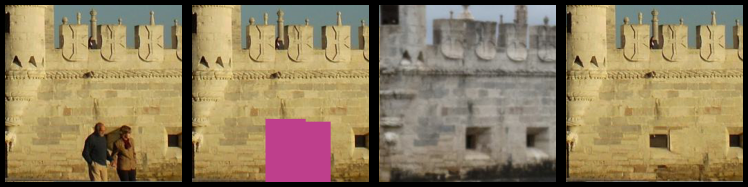}\\
    \caption{ 3D-consistent inpainting with \name: a) original image, b) image with masked occlusions; c) in-context image; d) \name \ inpainting preserving 3D consistency.}    
    \label{fig:teaser}
\end{figure}
%---------------------------------------------
%-------------------------------------------------------
 
Despite their impressive performance in generating photo-realistic 2D images, diffusion models still lack the 3D understanding during the generation process
\cite{anciukevicius23renderdiffusion,zhan2024generalprotocolprobelarge,huang2024diffusion_editing_survey}. Their inability to control the 3D properties of objects in the generated images
is particularly harmful for % negatively %hurts  %even more %conditional 
image inpainting, which aims to harmonizing 3D properties of masked and non-masked regions. 
3D understanding of diffusion models has been recently addressed %in unconditional DMs 
by adding explicit 3D priors~\cite{huang2024diffusion_editing_survey,li23generalised} 
or incorporating additional points of view~\cite{anciukevicius23renderdiffusion}; 
however these methods require 3D % or depth 
supervision which is hard to obtain~\cite{zhang23_coherent_image_inpainting}.

In this paper, we address the problem of {\it 3D-consistency of image inpainting} and propose {\it in-context inpainting} by injecting additional information into the image generative model, {\it without} 3D supervision. We train a generative model with image pairs that show the same scene and provide complementary information on the scene geometry, possibly under different lighting conditions. We exploit this complementarity %multi-view information 
when a part of the scene is occluded in one image but observed in another. %others % frames
At inpainting step, the model receives an image with a mask and inpaints the masked part while preserving the 3D scene information.

To achieve the 3D-consistent inpainting, we modify the generative diffusion model. 
We train our model %to denoise 2D images. %Our key insight is 
by incorporating the additional point of view of the scene into the denoising process. 
This creates an inductive bias that allows to recover 3D priors while training to denoise in 2D, without explicit 3D supervision. Training unconditional diffusion models with additional images as in-context guidance allows us to harmonize the masked regions while inpainting and ensuring the 3D consistency.
We evaluate our method, named \name, on one synthetic (HM3D) and three real-world  (MegaDepth, StreetView and WalkingTour) datasets and show that it generates %plasusible harmonized or smooth
semantically harmonized and 3D-consistent inpaintings outperforming the state-of-the-art methods. Figure~\ref{fig:teaser} shows examples of inpainting with \name, 
with occlusions caused by the presence of cars, buses and pedestrians.

Our contributions can be summarized as follows:
\begin{itemize}
\item 
We train an unconditional diffusion model with an additional image as the in-context guidance 
for image generation. 
\item 
We apply post-conditional generative approach to image inpainting where the mask on first image is given at the inference time; we modify the resampling schedule for harmonizing boundaries between masked an non-masked regions of the image.  
\item 
We prove its efficiency of \name \ for image inpainting with semantic and random masks on one synthetic and three real-world datasets.
\end{itemize}

\section{Related Work}
\label{sec:related_work}
\noindent

\PAR{Image Inpainting}
%The common set-up for %In the nutshell, 
The task of image inpainting inputs an image and a binary mask that defines the erased pixels \cite{yu2019free, liu2022partial}. The original image pixels are removed and new ones are generated based on the mask. The domain of image inpainting was previously dominated by Generative Adversarial Networks (GANs)~\cite{li2020recurrent,pathak2016context,liu2018image}
that use an encoder-decoder architecture as the main inpainting generator, adversarial training, and tailored losses that aim at photo-realism~\cite{liu2022partial,yildirim2023diverse}. Most GAN-based models output deterministic results as these models are trained with reconstruction losses and improved stability~\cite{yu2019free,yu2022high}.
Recent GAN-based methods achieve diversity~\cite{zhao2021large, li2022mat};
however, trained on single object datasets, 
they are not extended to inpaint other scenes.
 
%---------------------------------------------
\PAR{Image Conditional Diffusion Models} 
Recent advances in diffusion models and the high expressiveness of pretrained Denoising Diffusion Probabilistic Models~\cite{ho20_denoising_diffusion,improvedddpm} led to using them as a prior for generic image inpainting. 
SohlDickstein~\etal~\cite{diffusionThermo} applied early diffusion models to inpainting. %Song~\cite{ddim}... 
Song~\etal~\cite{sde} proposed a score-based formulation using stochastic differential equations for unconditional image generation, with an additional application to inpainting. Score-based diffusion models was applied to conditional image generation in \cite{Batzolis21conditional} and extended it to multi-speed diffusion.
Ho~\etal~\cite{ho2022classifier} jointly train a conditional and an unconditional diffusion models and combine the resulting conditional and unconditional score estimates.
Image-to-image translation with diffusion models was proposed
in~\cite{saharia22palette}, with an application to inpainting. 

%---------------------------------------------
\PAR{Unconditional Diffusion Models}
RePaint~\cite{lugmayr22repaint} was first to show that a pretrained unconditional diffusion model can inpaint images without mask-conditioned training. 
It modifies the denoising process to condition generation on the non-masked image content. Similarly, X-Decoder~\cite{zou2022generalized} processes an input image and textual together for referring segmentation. It proceeds to prompt-based segmentation and, when combined with diffusion models, can erase the segmented objects. %operating in pixel~\cite{lugmayr22repaint} 
LatentPaint~\cite{Corneanu2024latentpaint} extended the denoising process from pixel to latent space. Trained and evaluated mostly on human face datasets, these models perform well on 2D inpainting however they perform poorly on street view images or indoor images with important scene geometry. In this paper, we are first to address 3D consistent image inpainting and to make an additional step by integrating additional images in the training unconditional DDPMs. 

%%%%%%%%%%%%%%%%%%%%%%%%%%%%%%%%%%%%%%%%%%%%%%%%%%%%%%%%%
\section{Diffusion Models for Image Inpainting}
\label{sec:method}
\noindent
We first present an unconditional DDPM and describe injecting in-context images in the reverse diffusion process. Then, we introduce 3D-consistent harmonization of masked and non-masked regions and modify the resampling schedule to accelerate the reverse process for the mask-based inpainting. 
\begin{figure} 
\centering
\includegraphics[width=\columnwidth]{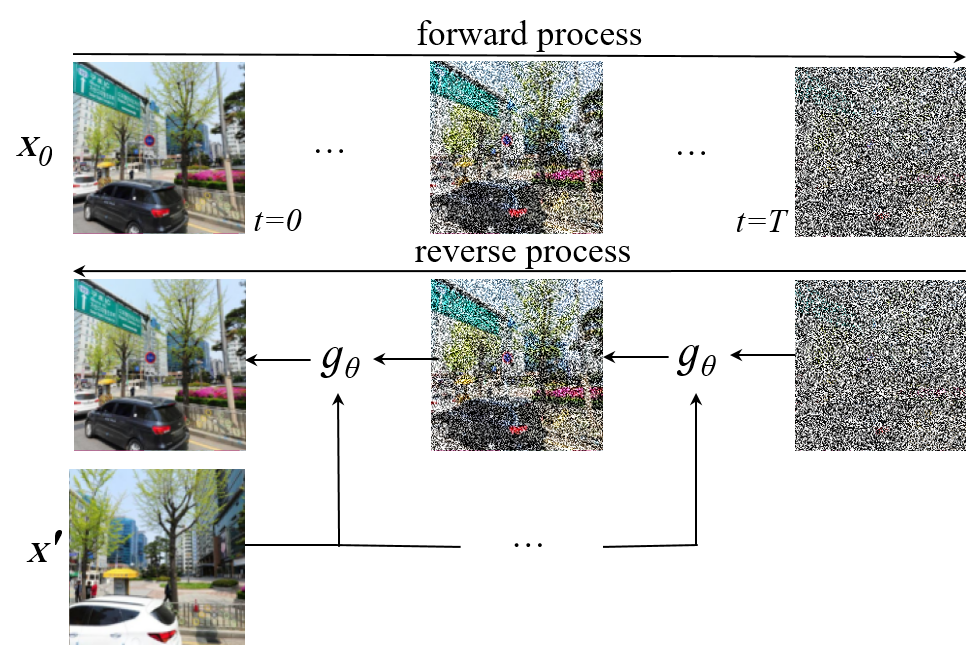}
\caption{Unconditional DDPM with the forward process (right arrows) and the reverse denoising process (left arrows) that takes into account additional image $\mathbf{x}'$.
}
\label{fig:diffusion}
\end{figure}

%=================================================
\subsection{DDPMs}
\label{ssec:image_diffusion}
Diffusion models are designed to generate an image $\mathbf{x}_0$ by moving a starting full noise image $\mathbf{x}_T \sim \mathcal{N}(\mathbf{0}, \mathbf{I})$ progressively closer to the data distribution through multiple denoising steps $\mathbf{x}_{T-1}, \dots, \mathbf{x}_0$. The models are divided into forward and reverse diffusion processes. 
During the forward process, noisy images $\mathbf{x}_1, \dots, \mathbf{x}_T$ are created by repeatedly adding Gaussian noise starting from a training image $\mathbf{x}_0$:
\begin{equation}
    q ( \mathbf{x}_t | \mathbf{x}_{t-1} ) \sim \mathcal{N}(\mathbf{x}_t; \sqrt{1-\beta_t}\mathbf{x}_{t-1}, \beta_t \mathbf{I}),
\label{eq:forward}
\end{equation}    
where $\beta_t$ is a variance schedule that increases from $\beta_0=0$ to $\beta_T=1$ and controls how much noise is added at each step. We initially use a cosine schedule~\cite{nichol2021glide} in our experiments and consider replacing it with alternative schedules in Section~\ref{ssec:resampling}.
The iterative process in (\ref{eq:forward}) can be reshaped to directly obtain $\mathbf{x}_t$ from $\mathbf{x}_0$ in a single step:
\begin{gather}
     q(\mathbf{x}_t | \mathbf{x}_0) = \sqrt{\bar{\alpha_t}}\mathbf{x}_{0} + \sqrt{1-\bar{\alpha_t}}\boldsymbol{\epsilon} \nonumber\\
     \text{with } \boldsymbol{\epsilon} \sim \mathcal{N}(0, \mathbf{I}), \bar{\alpha_t} 
     \coloneqq \alpha_1 \ldots \alpha_t \text{ and } \alpha_t \coloneqq (1-\beta_t). 
     \label{eq:forward_noise}
\end{gather}
The reverse process aims at rolling back the steps of the forward process by finding the posterior distribution for the less noisy image $\mathbf{x}_{t-1}$ given the more noisy image $\mathbf{x}_t$:
\begin{align}
    q(\mathbf{x}_{t-1} | \mathbf{x}_t, \mathbf{x}_0) & \sim \mathcal{N}(\mathbf{x}_{t-1}; \boldsymbol{\mu_t}, \sigma_t^2\mathbf{I}), \\
    \text{where~ } \boldsymbol{\mu_t} &\coloneqq \frac{\sqrt{\bar{\alpha}_{t-1}}\beta_t}{1-\bar{\alpha}_{t}} \mathbf{x}_0 + \frac{\sqrt{\alpha_t}(1-\bar{\alpha}_{t-1})}{1-\bar{\alpha}_{t}}\mathbf{x}_t \nonumber \\
    \text{and~ } \sigma_t^2 &\coloneqq \frac{1-\bar{\alpha}_{t-1}}{1-\bar{\alpha}_t}\beta_t. \nonumber 
    \label{eq:backward}
\end{align}
The image $\mathbf{x}_0$ being unknown, the distribution $q$ can not be directly computed.
Instead, DMs train a neural network $g$ with parameters $\theta$ to approximate $q$
and predicts the parameters $\mu_{\theta}(\mathbf{x}_t, t)$ and $\Sigma_{\theta}(\mathbf{x}_t, t)$ of a Gaussian distribution:
\begin{equation}
\label{eq:nn}
p_{\theta}(\mathbf{x}_{t-1}|\mathbf{x}_t) = \mathcal{N}(\mathbf{x}_{t-1}; \mu_{\theta}(\mathbf{x}_t, t), \Sigma_{\theta}(\mathbf{x}_t, t)).
\end{equation}
The learning objective for the model (\ref{eq:nn})  %(\eqref{eq:nn}) 
is derived by considering the variational lower bound of the density assigned to the data by the model~\cite{ho20_denoising_diffusion}.
This approximate posterior is sampled at each generation step to progressively get the less noisy image $\mathbf{x}_{t-1}$ from the more noisy image $\mathbf{x}_t$. 
It allows for a closed form expression of the objective since $q(\mathbf{x}_{t-1}|\mathbf{x}_t,\mathbf{x}_0)$ is also Gaussian. %~\cite{ddpm}.
 
Instead of predicting $\mathbf{x}_0$, the seminal work of Ho~\etal~\cite{ho20_denoising_diffusion} proposed to predict the cumulative noise $\epsilon$ 
that is added to the current intermediate image $\mathbf{x}_t$. We use the following parametrization of the predicted mean $\mu_\theta(\mathbf{x}_t, t)$:
\vspace{-3mm}
\begin{equation}
\vspace{-2mm}
\mu_{\theta}(\mathbf{x}_t, t) = \frac{1}{\sqrt{\alpha_t}} \left(\mathbf{x}_t - \frac{\beta_t}{\sqrt{1-\bar{\alpha}_t}} g_{\theta}(\mathbf{x}_t, t) \right)
\end{equation}
and derive the following training objective 
\begin{equation}
L_{\text{s}} = E_{t,\mathbf{x}_0,\epsilon} ||\epsilon - g_{\theta}(\mathbf{x}_t, t) ||^2.
\end{equation}

Beyond the predicted mean, learning the variance $\Sigma_{\theta}(\mathbf{x}_t, t)$ in (\ref{eq:nn}) of the reverse process reduces further the number of sampling steps and the inference time %by an order of magnitude 
~\cite{improvedddpm}.

%---------------------------------
\begin{figure} 
\centering
\includegraphics[width=\columnwidth]{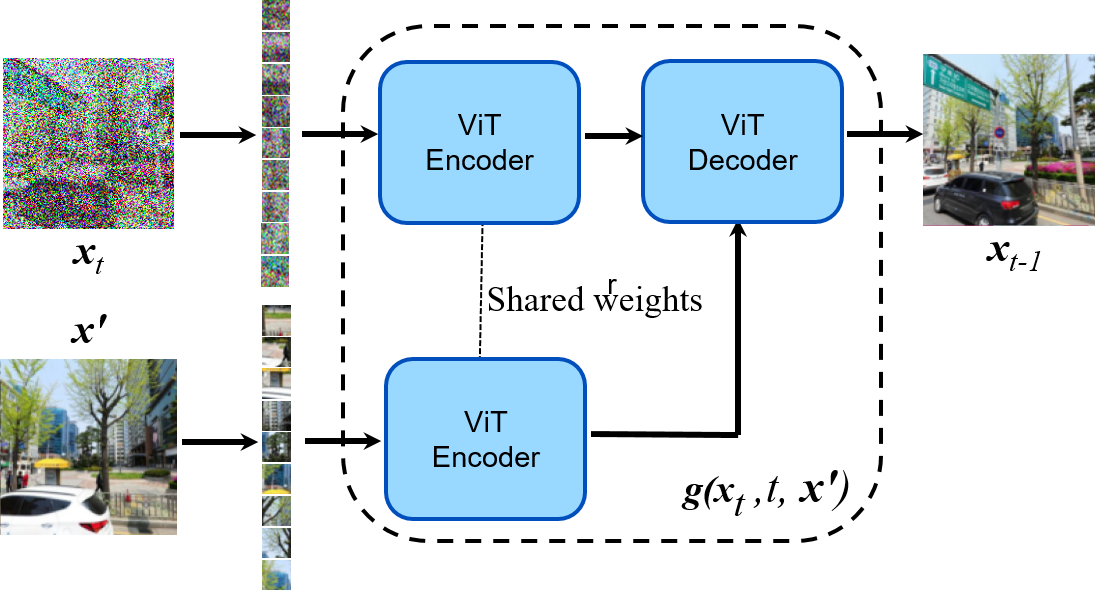}
\caption{ViT architecture for learning the denoising model with additional image $\mathbf{x}'$.}
\label{fig:vit}
\end{figure} 
%---------------------------------

%%%%%%%%%%%%%%%%%%%%%%%%%%%%%%%%%%%%%%%% 
\subsection{DDPMs with in-context images} 
\label{ssec:croco}
\noindent
We modify the training of unconditional DDPMs by injecting in-context images (see Figure~\ref{fig:diffusion}) to better approximate the target distribution $q$ and improve 3D consistency of image inpainting. 
Unlike the majority of generative diffusion models~\cite{ho20_denoising_diffusion,improvedddpm} and inpainting methods~\cite{lugmayr22repaint,yu2023inpaint}, which use U-Net backbones,
we adopt the class of diffusion models based on Diffusion Transformers (DiTs)~\cite{peebles23scalable} which adhere to the best practices of Vision Transformers (ViTs) and make injecting of in-context image $\mathbf{x}'$ straightforward.

By injecting the second image $\mathbf{x}'$, 
the objective function of our model is also to predict the cumulative noise  %$\epsilon_0$ 
$\epsilon$ that is added to the current intermediate image $\mathbf{x}_t$,  %follows:
\begin{equation}
L_{\text{2}} = E_{t,\mathbf{x}_0,\epsilon} ||\epsilon - g_{\theta}(\mathbf{x}_t, t, \mathbf{x}') ||^2.
\end{equation}

%%%%%%%%%%%%%%%%%%%%%%%%%%%%%%%%%%%%%%%% from croco-dvo
For  training the network $g_\theta(\mathbf{x}_t, t, \mathbf{x}')$, we adopt the principle of \textit{cross-view completion} which was shown to enable a network to perceive low-level geometric cues highly relevant to 3D vision tasks~\cite{croco2022,weinzaepfel2023croco}. 
The architecture of our network is shown in Figure~\ref{fig:vit}.
It takes an input the noised version of the first image ${\mathbf x}_t$ and the clean version of image ${\mathbf x}'$. Both images are patchified 
%\todo{masked, the masked patches are discarded}, % from Croco %and the remaining ones are 
and fed to an image encoder which is implemented using a ViT backbone. All patches from the second image are encoded using the same ViT encoder with shared weights. 
The latent token representations output by the encoder from both images  
%, including tokens to account for the masked patches of the first image, 
are then fed to a decoder whose goal is to %predict %the appearance patches. 
denoise ${\mathbf x}_t$. % the first (noised image.
The decoder uses a series of transformer decoder blocks comprising cross-attention layers. This allows 
noised tokens from the first image to attend clean tokens from the second image, thus enabling cross-view comparison and reasoning. The model is trained using a pixel reconstruction loss over all patches, similar to %MAE~\cite{assran2022masked} and 
CroCo~\cite{croco2022,weinzaepfel2023croco}. 

%=================================================================
\subsection{Conditioning DDPMs on the known region}
\label{ssec:resampling}
At the inference step, we predict missing regions of an image defined by a mask region $m$; 
we use the mask to condition the generative DDPM presented in the previous section.  
We first use DINOv2~\cite{oquab2024dinov2learningrobustvisual} for detecting and masking the occlusion classes like e.g. {\it vehicle} and {\it pedestrian}; 
we then mask any patch containing at least 1 masked pixel. 
We follow~\cite{lugmayr22repaint} and denote  
the masked regions as $m \odot {\mathbf x}^m$ and the known (non-masked) regions as $(1 - m) \odot {\mathbf x}^k$. Since every reverse step %(2) -- > (2), (7) --> (3) ?
(\ref{eq:nn}) from ${\mathbf x}_t$ to ${\mathbf x}_{t-1}$ depends on ${\mathbf x}_t$, we can alter the known regions $(1-m) \odot {\mathbf x}_t$ as long as we keep the correct properties of the 
target distribution. Since the forward process is defined by %a Markov Chain 
(\ref{eq:forward}) as accumulative Gaussian noise, we can sample the intermediate image ${\mathbf x}_t$ at any point in time using (\ref{eq:nn}). % (7).
This allows us to sample the known regions $m \odot {\mathbf x}_t$ at any time step $t$. Using (\ref{eq:forward_noise}) and (\ref{eq:nn}) for the masked and known regions, respectively, one reverse step can be described by the following expression,  
\begin{equation}
\begin{tabular}{rlll}
${\mathbf x}^{k}_{t-1}$ &$\approx$ & $N (\sqrt{\hat \alpha_t} {\mathbf x}_0, (1-\alpha_t){\bf I})$ & {\it known regions} \\
${\mathbf x}^{m}_{t-1}$ &$\approx$ & $N (\mu_{\theta} (x_t, t), \Sigma_{\theta} (x_t, t))$         & {\it masked regions} \\
${\mathbf x}_{t-1}$    &   =      & $m \odot {\mathbf x}^{m}_{t-1} + (1-m)\odot {\mathbf x}^{k}_{t-1}$ & {\it combined} \\ 
\end{tabular}
\label{eq:2}
\end{equation}
Therefore, ${\mathbf x}^{k}_{t-1}$ is sampled using the known regions in the image $m \odot {\mathbf x}_0$, while ${\mathbf x}^{m}_{t-1}$ is sampled from the model, given the previous iteration ${\mathbf x}_t$. Then they are combined to the new sample ${\mathbf x}_{t-1}$. % using 

%-------------------------------------------------------- ----
\begin{figure*}
\centering
\newcommand{\figpairwidth}{0.115\linewidth}
\resizebox{\linewidth}{!}{
\begin{tabular}{@{}l@{~ }p{\linewidth}@{}}
\rotatebox{90}{\small HM3D} & 
\includegraphics[width=\figpairwidth]{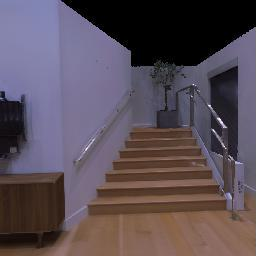}
\includegraphics[width=\figpairwidth]{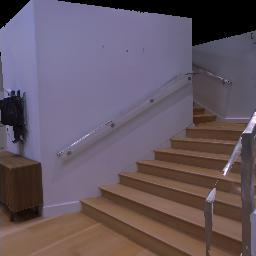} \hfill
\includegraphics[width=\figpairwidth]{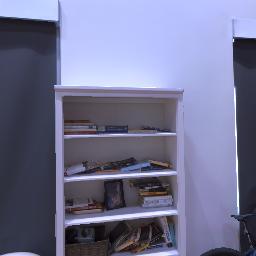}
\includegraphics[width=\figpairwidth]{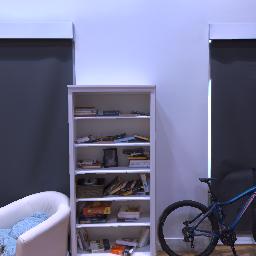} \hfill
\includegraphics[width=\figpairwidth]{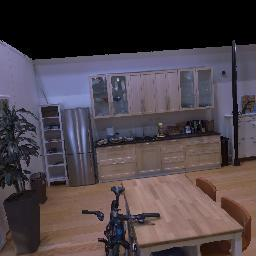}
\includegraphics[width=\figpairwidth]{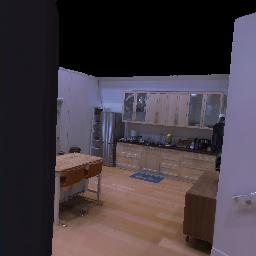} \hfill
\includegraphics[width=\figpairwidth]{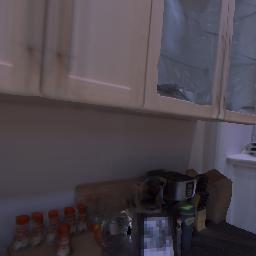}
\includegraphics[width=\figpairwidth]{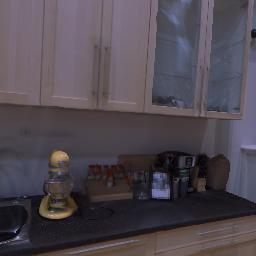} \hfill
\\[-0.1cm]
\rotatebox{90}{\small ~~~MegaDepth} & 
\includegraphics[width=\figpairwidth]{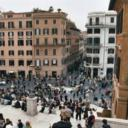}
\includegraphics[width=\figpairwidth]{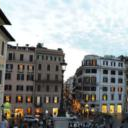} \hfill
\includegraphics[width=\figpairwidth]{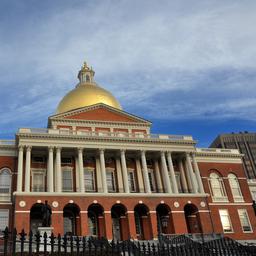}
\includegraphics[width=\figpairwidth]{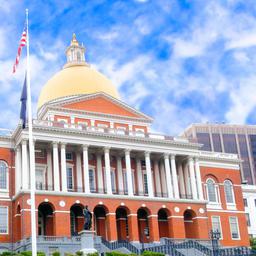} \hfill
\includegraphics[width=\figpairwidth]{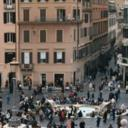}
\includegraphics[width=\figpairwidth]{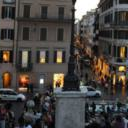} \hfill
\includegraphics[width=\figpairwidth]{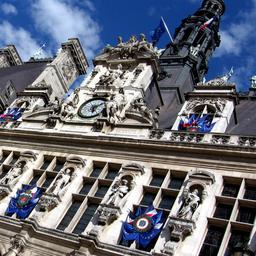}
\includegraphics[width=\figpairwidth]{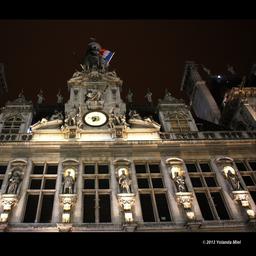} \hfill
\\[-0.1cm]
\rotatebox{90}{\small ~StreetView} &
\includegraphics[width=\figpairwidth]{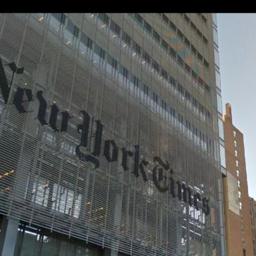}
\includegraphics[width=\figpairwidth]{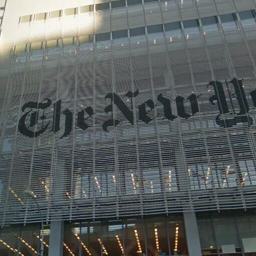} \hfill
\includegraphics[width=\figpairwidth]{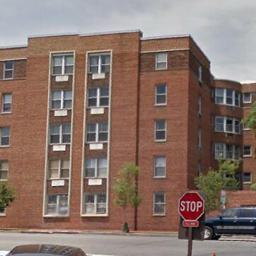}
\includegraphics[width=\figpairwidth]{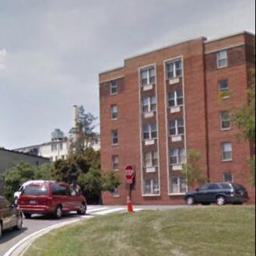} \hfill
\includegraphics[width=\figpairwidth]{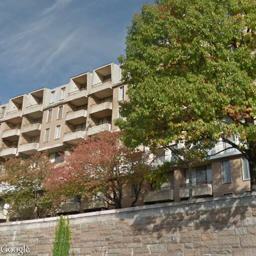}
\includegraphics[width=\figpairwidth]{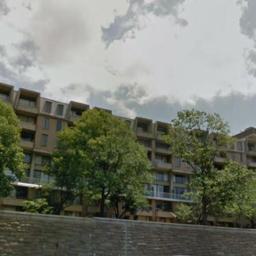} \hfill
\includegraphics[width=\figpairwidth]{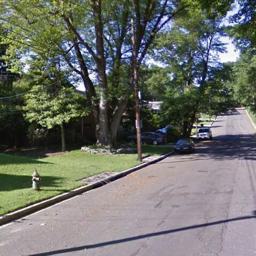}
\includegraphics[width=\figpairwidth]{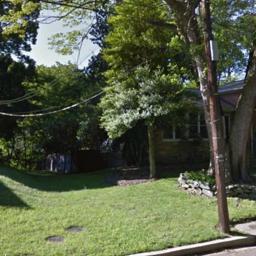} \\
[-0.1cm]
\rotatebox{90}{\small ~Amsterdam} &
\includegraphics[width=\figpairwidth]{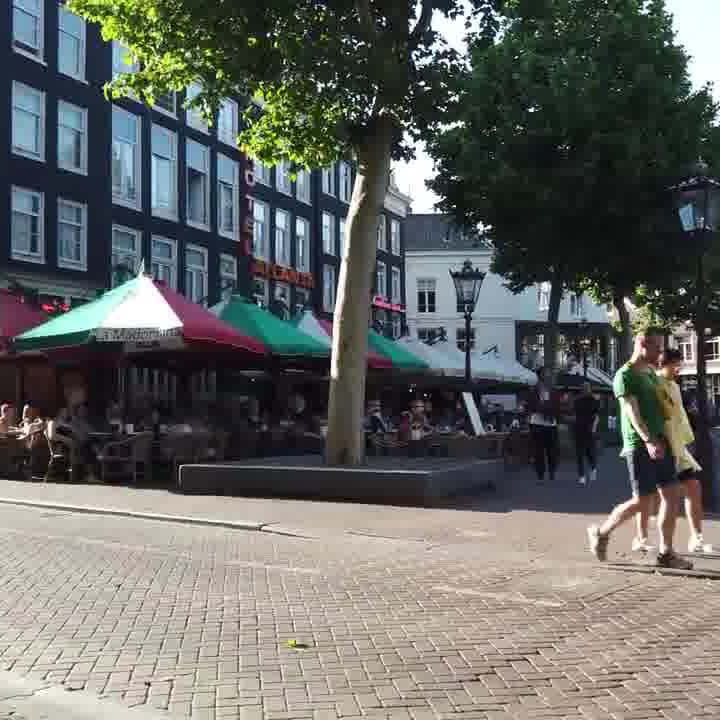}
\includegraphics[width=\figpairwidth]{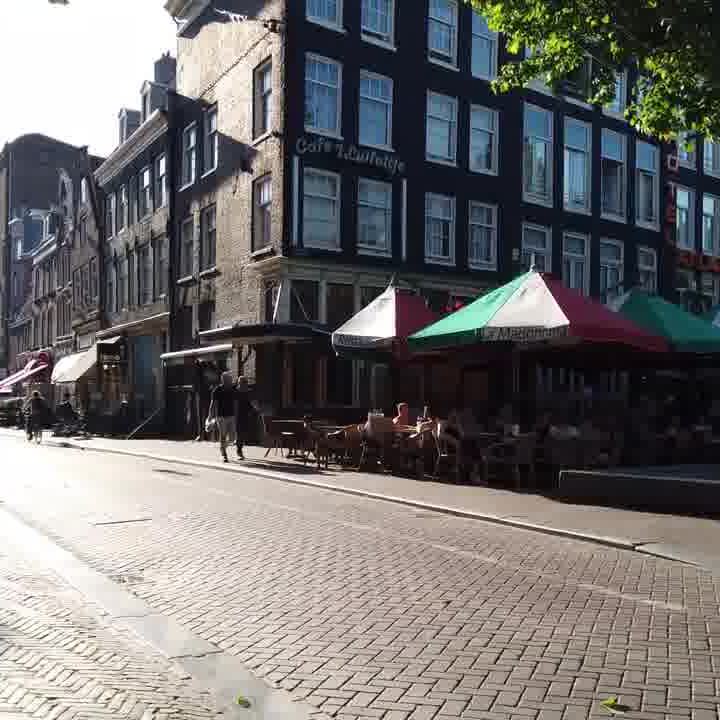} \hfill
\includegraphics[width=\figpairwidth]{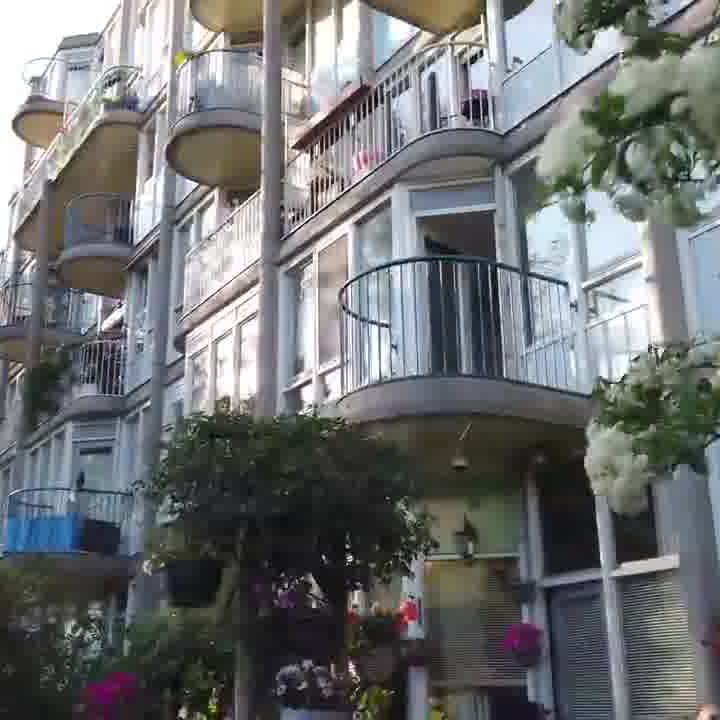}
\includegraphics[width=\figpairwidth]{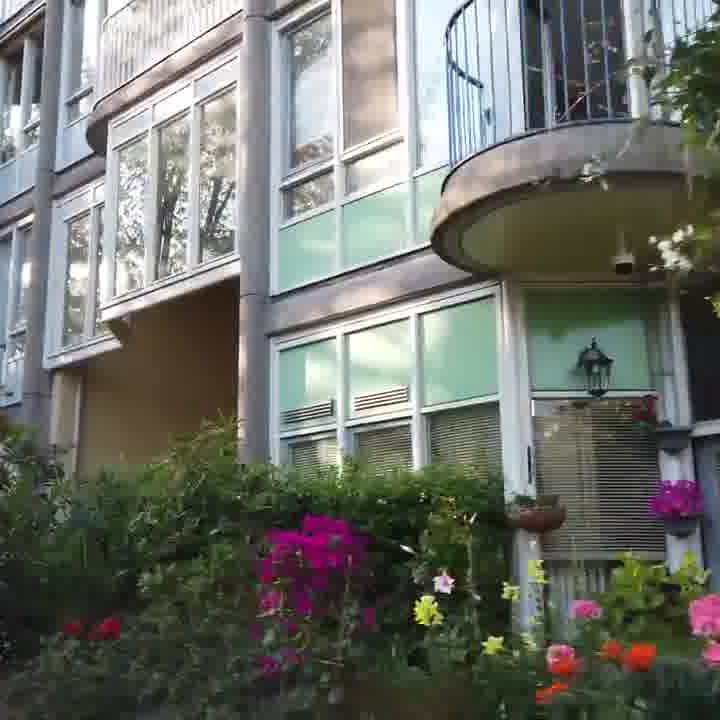} \hfill
\includegraphics[width=\figpairwidth]{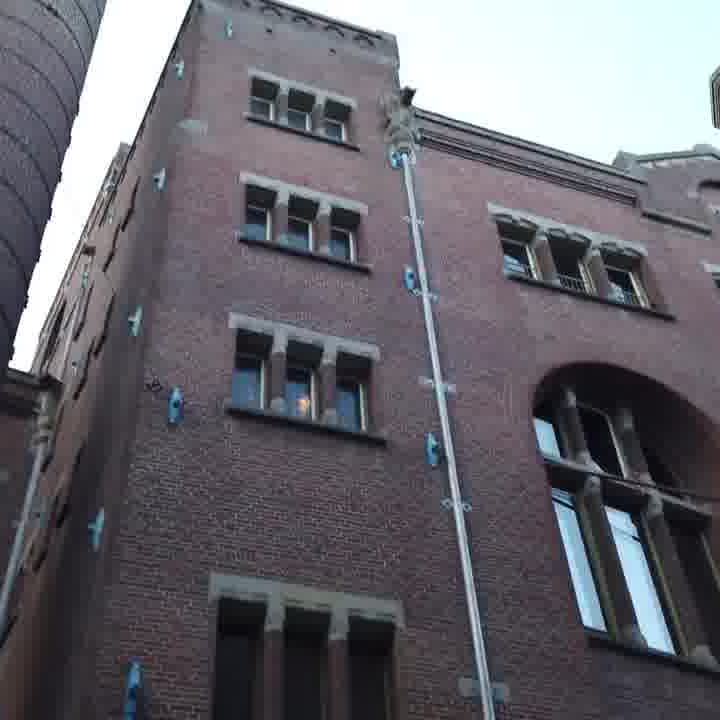}
\includegraphics[width=\figpairwidth]{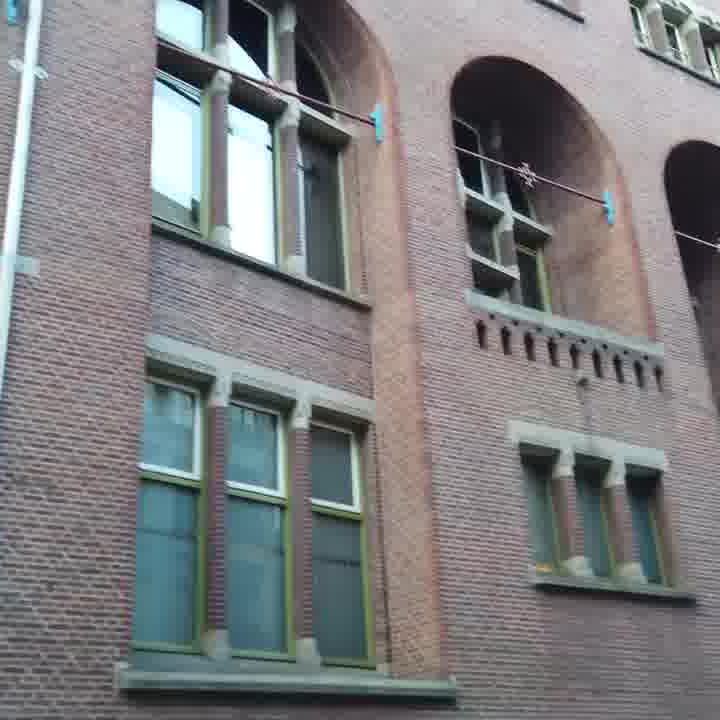} \hfill
\includegraphics[width=\figpairwidth]{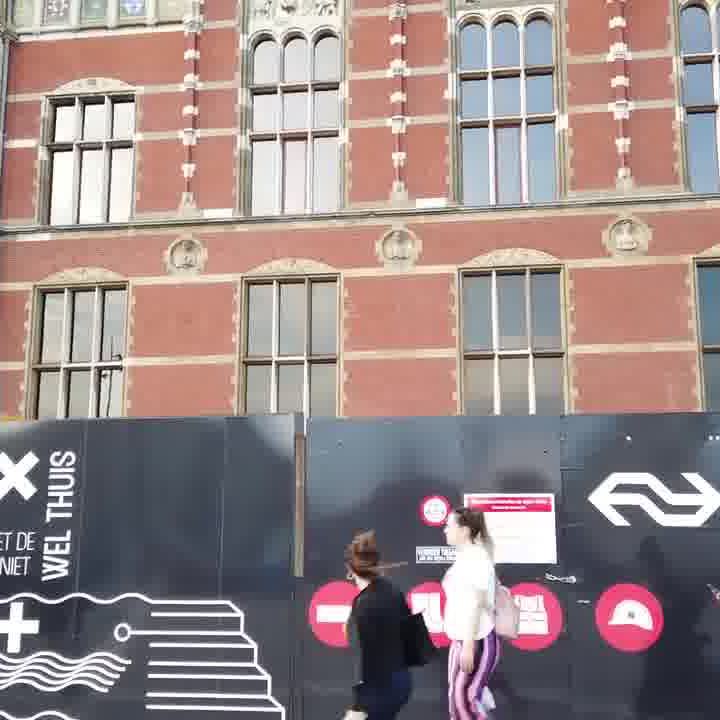}
\includegraphics[width=\figpairwidth]{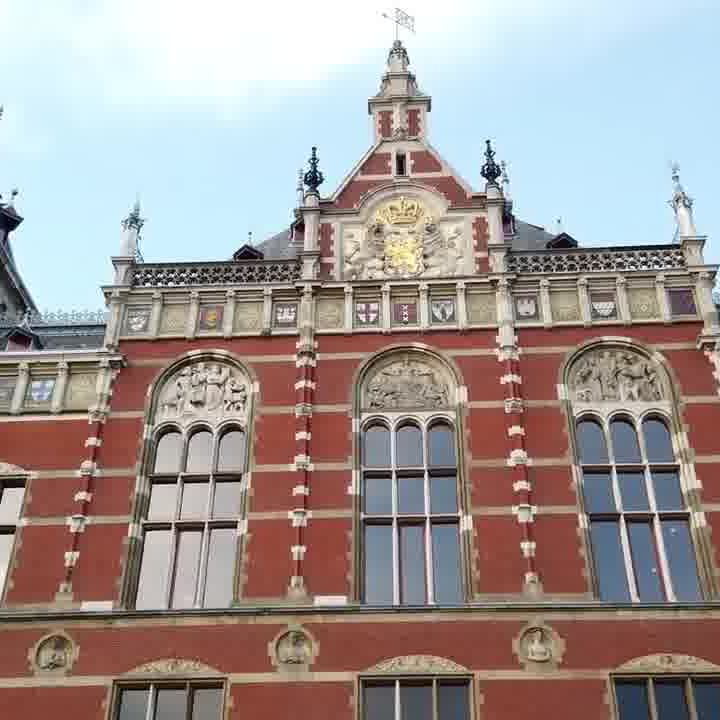} \hfill \\
[-0.1cm]
\rotatebox{90}{\small ~Singapour} &
\includegraphics[width=\figpairwidth]{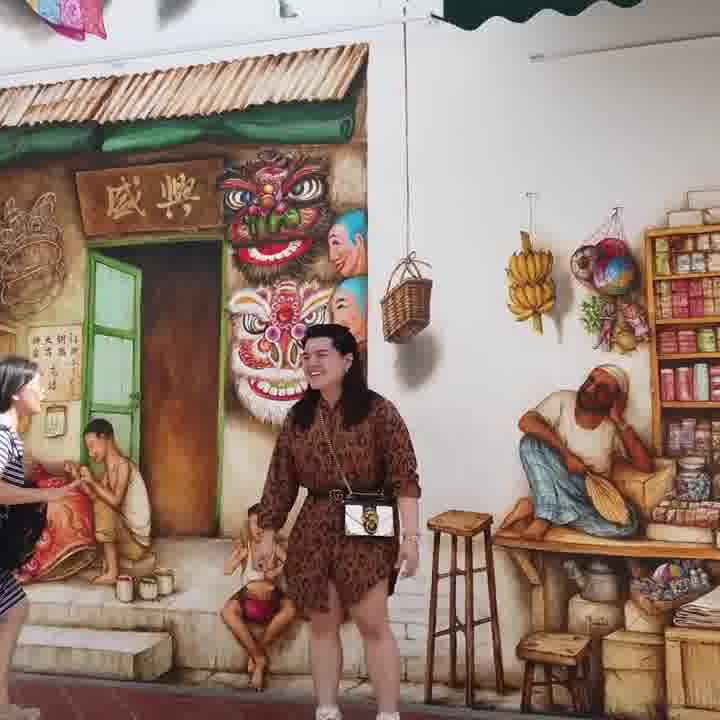}
\includegraphics[width=\figpairwidth]{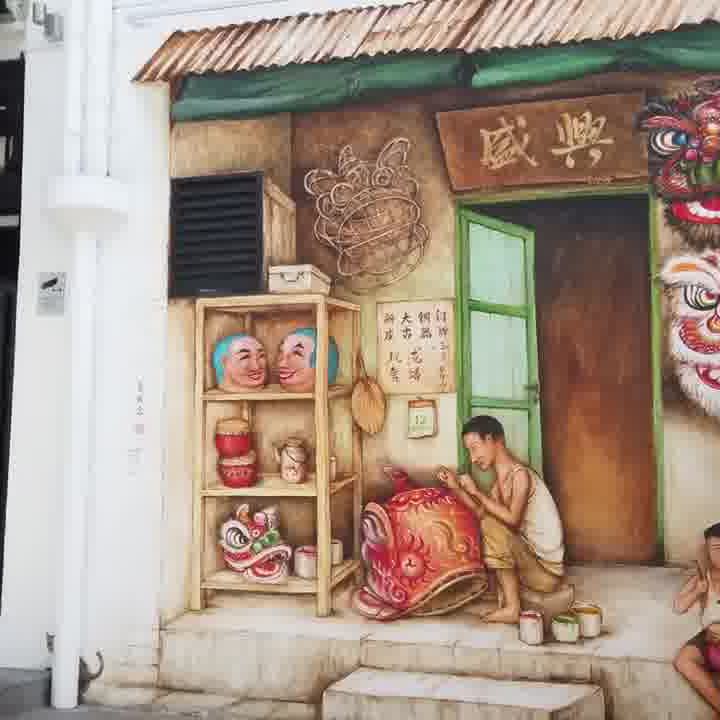} \hfill
\includegraphics[width=\figpairwidth]{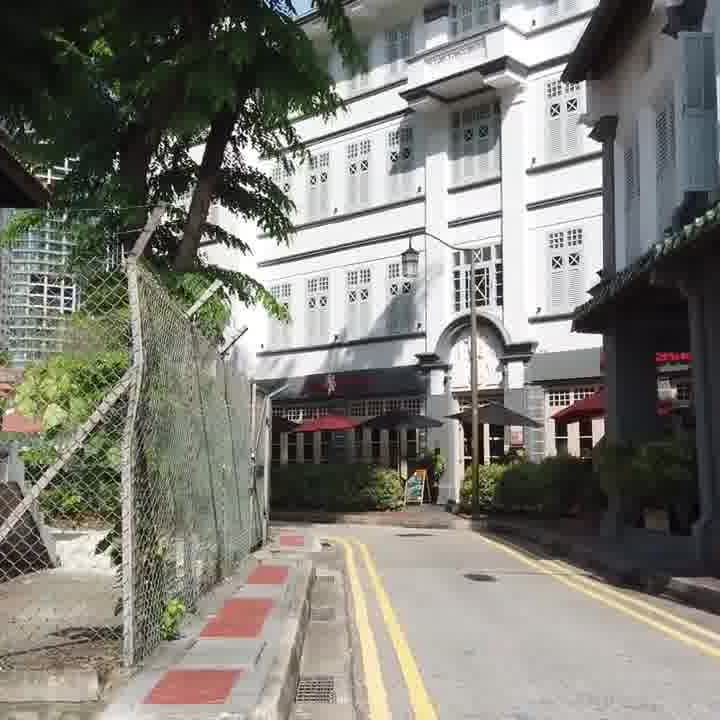}
\includegraphics[width=\figpairwidth]{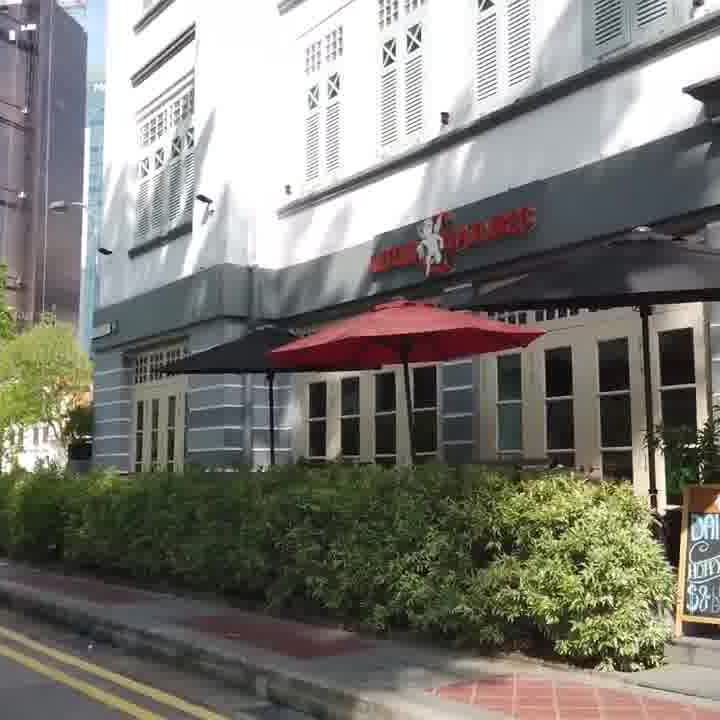} \hfill
\includegraphics[width=\figpairwidth]{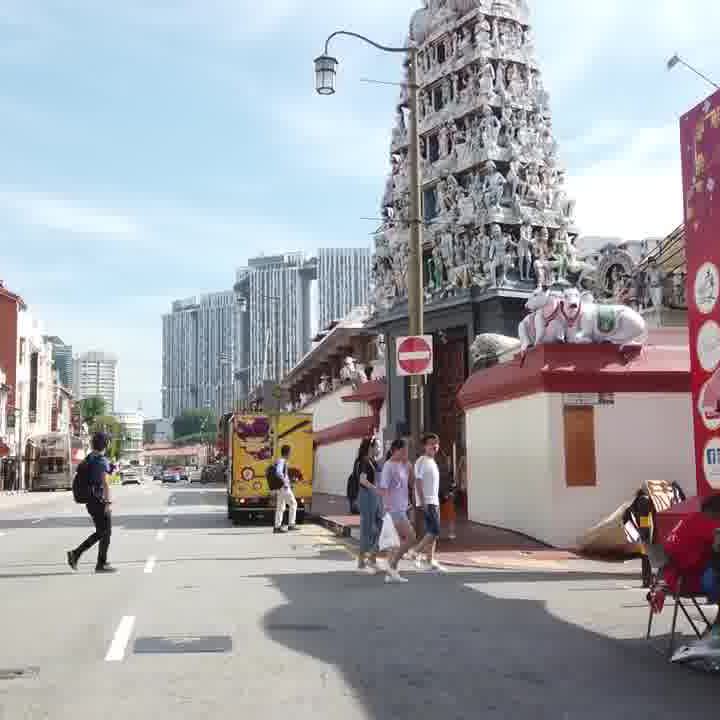}
\includegraphics[width=\figpairwidth]{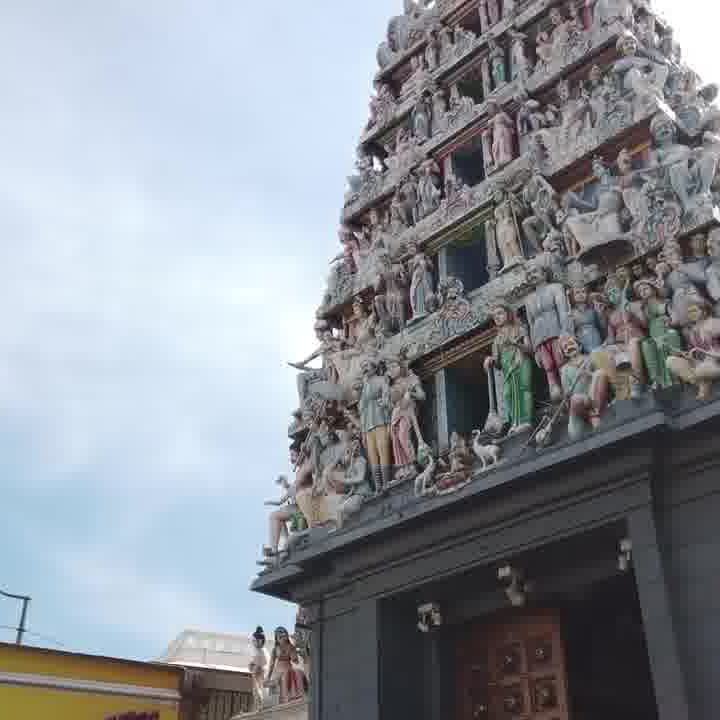} \hfill
\includegraphics[width=\figpairwidth]{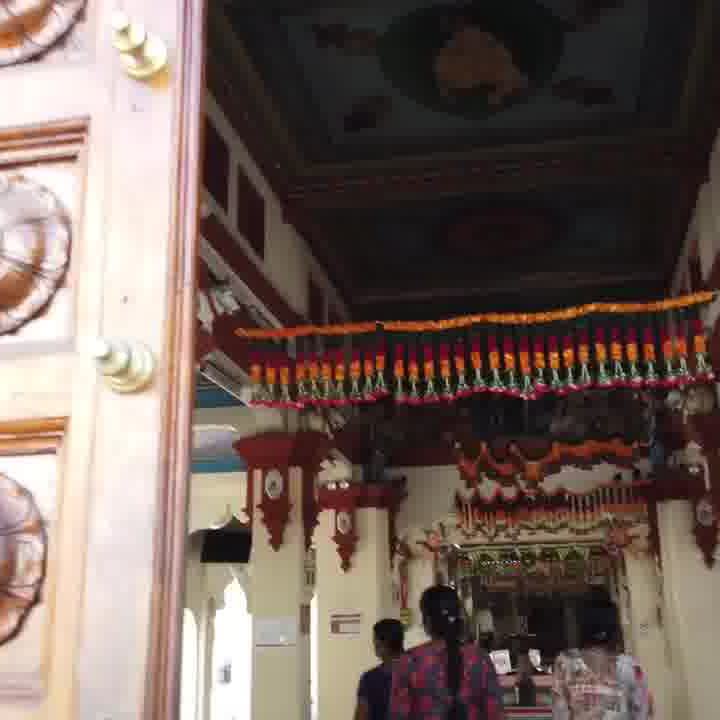}
\includegraphics[width=\figpairwidth]{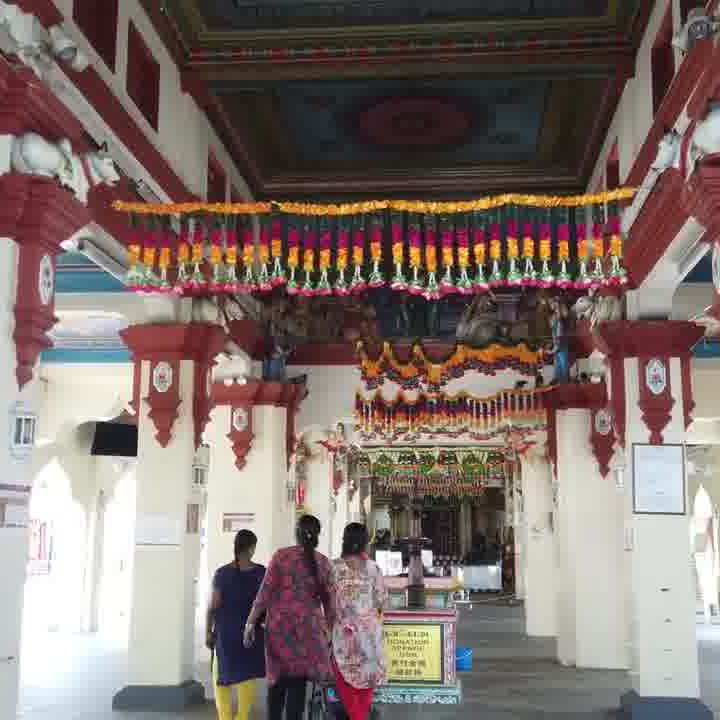} \hfill \\
[-0.1cm]
\end{tabular}
}
\caption{Image pairs for training unconditional DDPMs from HM3D, MegaDepth, StreetView and WalkingTour datasets.}
\label{fig:pairs}
\vspace{-0.3cm}
\end{figure*}
%-------------------------------------------------------------

\begin{figure*}
\centering
\setlength{\tabcolsep}{0.5pt}
\newcommand{\figpairwidth}{0.12\linewidth}
\resizebox{\linewidth}{!}{
\begin{tabular}{cccccccccc}
Source &Masked &In-context &Schedule &$t$=1000 &$t$=800 &$t$=790 &$t$=650 &$t$=640 &$t$=600 \\
\includegraphics[width=\figpairwidth]{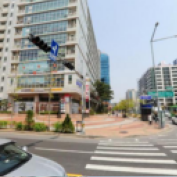} &
\includegraphics[width=\figpairwidth]{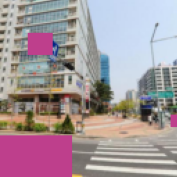} &
\includegraphics[width=\figpairwidth]{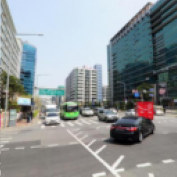} &
\includegraphics[width=\figpairwidth]{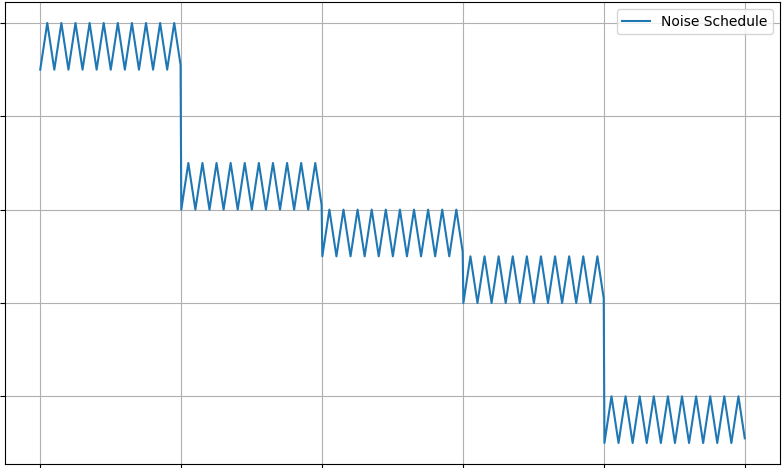} &
\includegraphics[width=\figpairwidth]{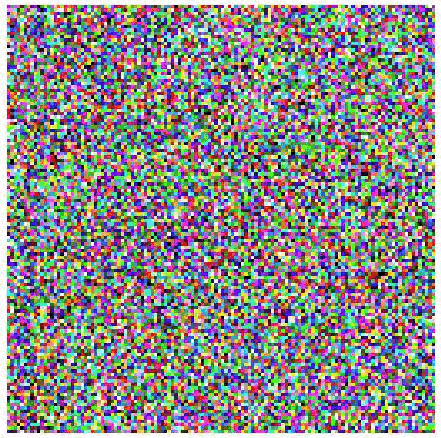} &
\includegraphics[width=\figpairwidth]{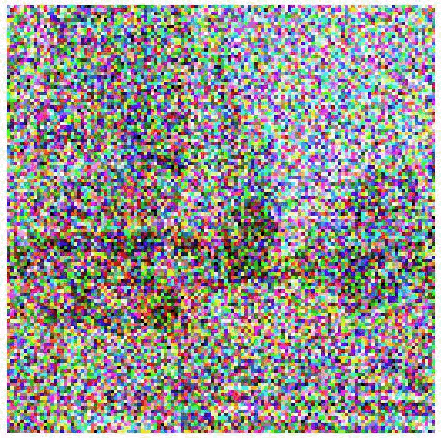} &
\includegraphics[width=\figpairwidth]{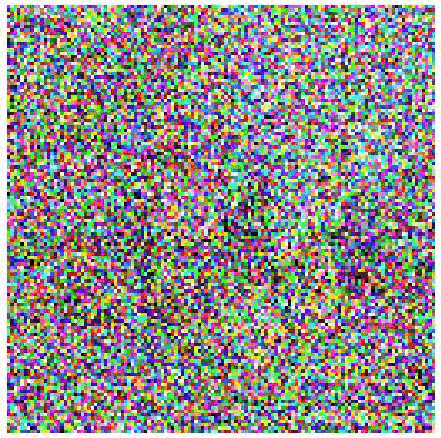} &
\includegraphics[width=\figpairwidth]{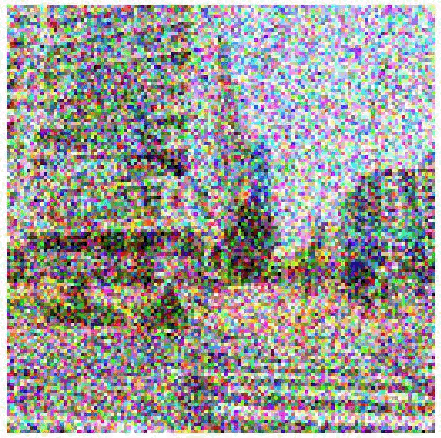} &
\includegraphics[width=\figpairwidth]{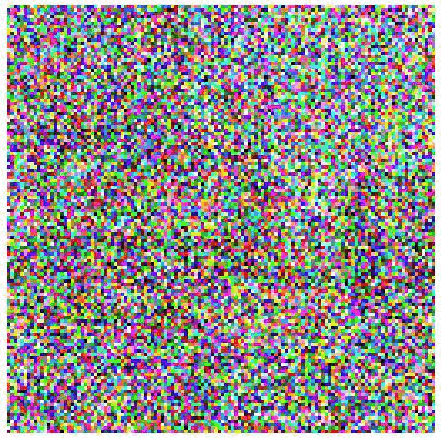} & 
\includegraphics[width=\figpairwidth]{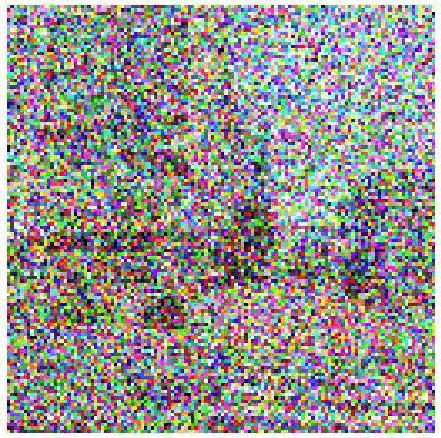}  \\ 
[-0.1cm]
$t$=540 &$t$=530 &$t$=490 &$t$=480 &$t$=300 &$t$=290 &$t$=100 &$t$=50 &$t$=40 &$t$=0 \\ 
\includegraphics[width=\figpairwidth]{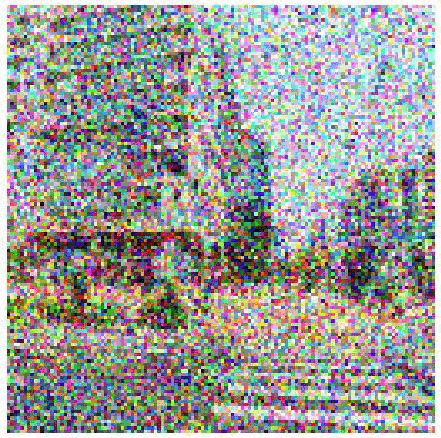} &
\includegraphics[width=\figpairwidth]{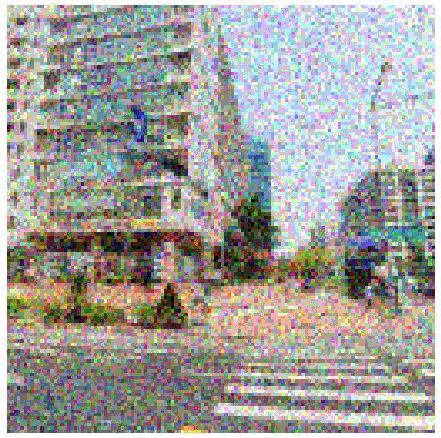} &
\includegraphics[width=\figpairwidth]{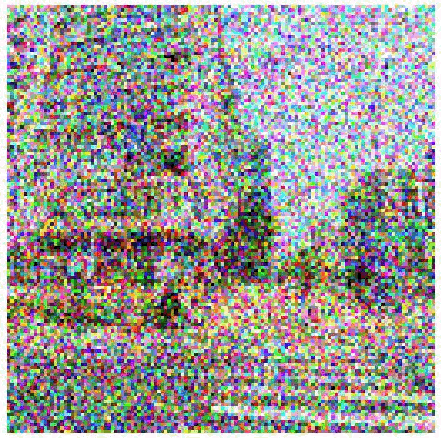} &
\includegraphics[width=\figpairwidth]{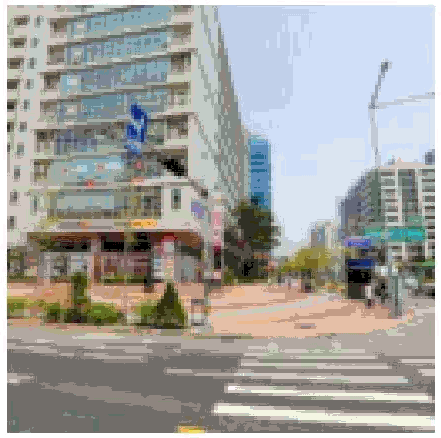} &
\includegraphics[width=\figpairwidth]{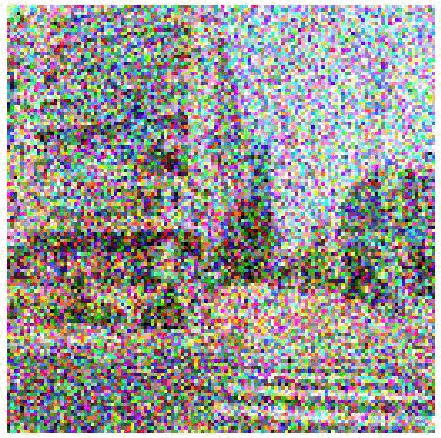} &
\includegraphics[width=\figpairwidth]{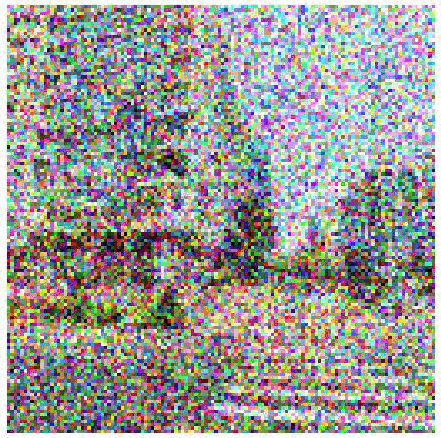} &
\includegraphics[width=\figpairwidth]{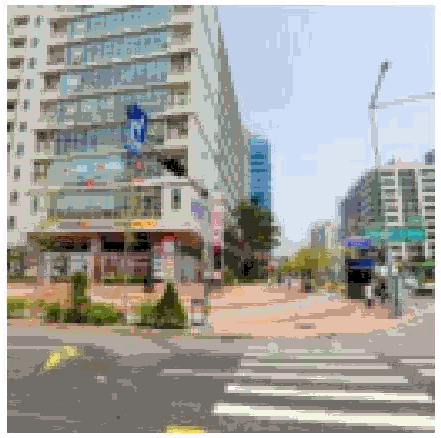} &
\includegraphics[width=\figpairwidth]{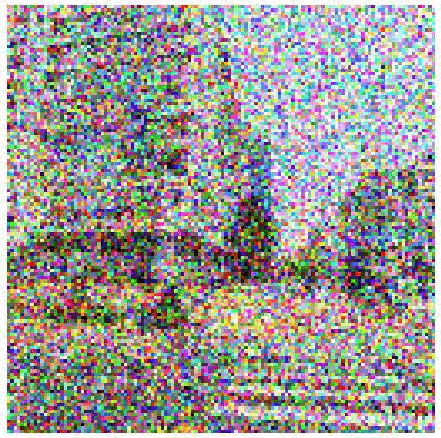} &
\includegraphics[width=\figpairwidth]{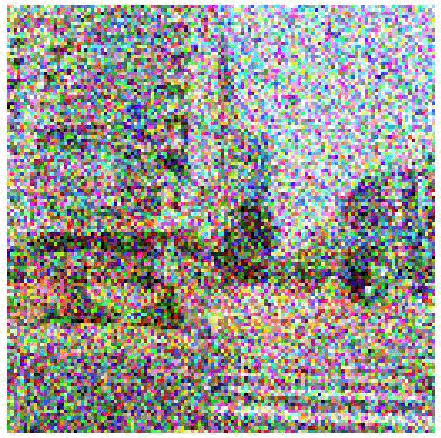} & 
\includegraphics[width=\figpairwidth]{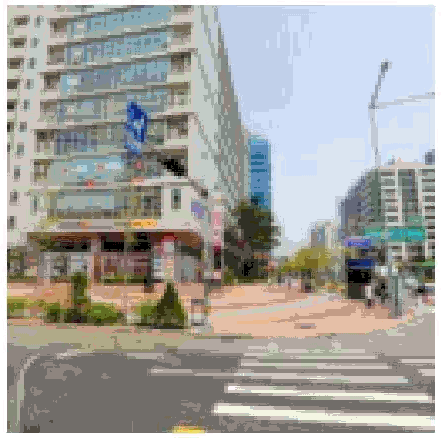}  \\ 
[-0.1cm]
\end{tabular}
}
\caption{Inpainting with semantic mask using noise schedule with jumps. Note resampling effect for harmonization.
%during the inpainting process.
}\label{fig:inference}
\vspace{-0.3cm}
\end{figure*}

Unfortunately, the basic denoising schedule (see Section~\ref{ssec:image_diffusion}) is insufficient for harmonizing the boundaries between masked and non-masked regions, due to restricted flexibility of sampling from both parts.
RePaint~\cite{lugmayr22repaint} paid a particular attention to the noise schedule and shown than the model needs more time to harmonize the conditional information ${\mathbf x}^{k}_{t-1}$ with the generated information ${\mathbf x}^{m}_{t-1}$ %at in one step 
before advancing to the next denoising step $t$.
They introduced a {\it resampling} approach %, we use this DDPM property 
to harmonize the masked and non-masked input of the model. %Consequently, we 
At an intermediate step, the resampling diffuses the output ${\mathbf x}_{t-1}$ back to ${\mathbf x}_t$ by sampling %from (ref{eq:} (1) as 
${\mathbf x}_t \approx N (\sqrt{1-\beta_t} x_{t-1}, \beta_t I)$.
By scaling back the output and adding noise,  %some 
information incorporated in the generated region ${\mathbf x}^{m}_{t-1}$ is still preserved in ${\mathbf x}^{m}_t$. It leads to a new ${\mathbf x}^{m}_t$ which is both more harmonized with ${\mathbf x}^{k}_t$ and contains conditional information from it. Figure~\ref{fig:inference} shows an example of inference process, with several phases of noise resampling jumps and diffusion. %and boundary harmonization.

Resampling with multiple jumps ~\cite{lugmayr22repaint} archives the boundary harmonization between known and masked segments, however it requires extra time for inference. 
To reduce the inference time, we consider alternative noise schedules. 
We additionally modulate the number of jumps over the entire denoising process by measuring the smoothness of the intermediate representations as proposed in~\cite{chen23importance}.

We modify the post-conditioned diffusion models by redefining the noise schedule, which is equivalent to importance sampling of the noise across different intensities. As demonstrated in \cite{hang2024improvednoiseschedulediffusion}, allocating more computation costs to intermediate noise levels yields superior performance compared to linearly increasing loss weights,
particularly under constrained computational budgets. By running experiments and analyzing the performance of different noise schedules, including Laplace, Cauchy, Cosine and their shifted and scaled versions~\cite{hang2024improvednoiseschedulediffusion}, we adopt the Laplace noise schedule which allows to reduce by half the overhead of multiple sampling from the distribution and the number of intermediate harmonization steps.

%%%%%%%%%%%%%%%%%%%%%%%%%%%%%%%%%%%%%%%%%%%%%%%%%%%%%%%%%
\section{Experiments}
\label{sec:evaluation}
\noindent
We first present one synthetic and three real-world datasets used in the experiments.
We also describe our approach to select image pairs suitable for training diffusion models with additional in-context images. 

$\circ \hspace{0.1cm} \textit{MegaDepth dataset}$ 
\cite{megadepth} consists of around 300K images downloaded from the web corresponding to 200 different landmarks. For each landmark, a point cloud model obtained using structure-form-motion (SfM) with COLMAP~\cite{colmapsfm} is also provided.

$\circ \hspace{0.1cm} \textit{Habitat-Matterport dataset (HM3D)}$
\cite{ramakrishnan2021hm3d} is the large dataset of synthetically generated 3D indoor spaces. It consists of 1K high-resolution 3D scans (or digital twins) of building-scale residential, commercial, and civic spaces generated from real-world environments, all available with 3D meshes and camera poses.

$\circ \hspace{0.1cm} \textit{StreetView dataset}$ contains 100K street view images from cities in South Korea, collected by Naver Maps\footnote{\tt https://map.naver.com/}. The images are provided with 3D location, camera pose and recording timestamps. 

$\circ \hspace{0.1cm} \textit{WalkingTour dataset}$ 
\footnote{\tt https://shashankvkt.github.io/dora/}
is a collection of egocentric videos captured in urban environments from cities in Europe and Asia. It consists of 10 high-resolution videos, each showcasing a person walking through a urban environment.

All the datasets offer ways of getting information about the geometry of the scene and the camera poses. To be useful for training in-context diffusion models and inpainting, we use the geometry information for each dataset %(point clouds, camera poses and 3D locations) 
to extract image pairs that depict a scene with some partial overlap. 
In addition to reasonable overlap, we ensure good diversity of selected pairs.
We follow~\cite{weinzaepfel2023croco} in obtaining an image pair quality score based on overlap and difference in viewpoint angle. Figure~\ref{fig:pairs} shows examples of image pairs extracted from all datasets. 

%-------------------------------------------------------------
%Example of mask-based inpainting with the modulated noise schedule.

%-------------------------------------------------------------

\PAR{Evaluation Metrics}
\label{ssec:metrics}
We use three common evaluation metrics to assess the performance and effectiveness of inpainting methods\cite{xiang23deep}.  

The {\it Peak Signal to Noise Ratio} (PSNR)~\cite{xia2018gibson} is a widely used simple metric for evaluating the quality of image inpainting and reconstruction. It measures the ratio between the maximum possible power of a signal and the power of the noise present in the signal. 

The {\it Learned Perceptual Image Patch Similarity} (LPIPS)~\cite{zhang2018unreasonableeffectivenessdeepfeatures} is a distance metric that is specifically designed to capture the perceptual similarity between two images. It is one of the most commonly used metrics for evaluating deep learning inpainting \& restoration systems. 

The {\it Structural Similarity Index} (SSIM) is a more advanced metric for evaluating the quality of image inpainting and reconstruction. It takes into account the human perception of image quality by measuring the structural similarity between two images, taking into account the luminance, contrast, and structure of the images. 

%----------------------------------------------------
\PAR{Implementation details}
Depth of the ViT encoder and decoder in \name \ is 12 and 8, the number of heads is 12 and 16, respectively~\cite{weinzaepfel2023croco}. 
For every dataset, all images are resized to $256 \times 256$; the patch size is 8. The network is trained on image pairs from scratch, using AdamW optimizer, the fixed learning rate 0.0004, for 1000 epochs. During the inference, for both semantic and random masks, we apply the resampling with 1000 steps and 10 jumps. 

%--------------------------------------
\subsection{Comparison to State-of-the-Art}
\label{ssec:comparsion}
We compare our \name\ method to the state of the art RePaint method
~\cite{lugmayr22repaint}. Due to unavailability of LatentPaint~\cite{Corneanu2024latentpaint} at the submission deadline, we complete the comparison with two powerful inpainting models, the Stable Diffusion %~\footnote{https://huggingface.co/docs/diffusers/api/pipelines/sd15}
\footnote{https://huggingface.co/runwayml/stable-diffusion-inpainting} and Stable Diffusion XL
\footnote{https://huggingface.co/diffusers/stable-diffusion-xl-1.0-inpainting-0.1}, known for excellent results on text- and mask-based inpainting of 2D images.

We note that RePaint~\cite{lugmayr22repaint} and LatentPaint~\cite{Corneanu2024latentpaint} are trained and evaluated primarily on human face datasets, and they underperform on datasets with important presence of 3D geometry, both indoor (HM3D) and outdoor (MegaDepth, StreetView, WalkingTour). 

We test our method with the {\it semantic mask} where obstacles are detected and masked as classes commonly representing obstacles such as {\it car}, {\it bus}, {\it bike}, {\it pedestrian}, {\it traffic light}, {\it pole}, etc. for outdoor scenes and {\it chair}, {\it armchair}, {\it table}, etc. for indoor scenes. Since considering certain class instances as obstacles (i.e. {\it table} for indoor, {\it pole} for outdoor scenes) is domain-specific and sometimes ambiguous, we also evaluate our method in a general case of {\it random mask}. Indeed, a randomly generated mask often covers multiple objects in different places of the scene, making the inpainting (both boundary harmonization and 3D consistency) harder. We proceed to the mask generation where a mask is a union of $n$=10 randomly selected rectangles, with the average masked ratio  of 0.4 of the image size.

%----------------------------------------------------
%----------------------------------------------------
\begin{table}[h] \centering
\setlength{\tabcolsep}{0.5pt}
\begin{tabular}{l|ccc|ccc}
\specialrule{1pt}{0.5pt}{0.2pt}
\rowcolor{GrayBorder}
Method   & \multicolumn{3}{c|}{Semantic Mask} & \multicolumn{3}{c}{Random Mask} \\ 
\rowcolor{GrayBorder}
         & PSNR$\uparrow$ & LPIPS$\downarrow$ & SSIM$\uparrow$ 
                               & PSNR$\uparrow$ & LPIPS$\downarrow$ & SSIM$\uparrow$ \\ \hline
\rowcolor{GrayBorder}
         & \multicolumn{6}{c}{MegaDepth} \\  
RePaint~\cite{lugmayr22repaint} 
       & 20.73 & 0.24 & 0.87             & 18.22 & 0.31 & 0.79  \\ 
SD     & 22.11 & 0.21 &{\bf 0.88}        & 22.96 & 0.19 & 0.84 \\
SD-XL  & 23.10 & 0.15 &{\bf 0.88}        & 23.14 & 0.15 & 0.84 \\  \hline
\name  &{\bf 23.39}&{\bf 0.12}&0.86      &{\bf 24.16}&{\bf 0.10}& {\bf 0.86} \\ 
\hline
\rowcolor{GrayBorder}
        & \multicolumn{6}{c}{StreetView} \\  
RePaint~\cite{lugmayr22repaint} 
       &17.92 & 0.30 & 0.81          & 17.16 & 0.34 & 0.78  \\ 
SD     &21.57 & 0.17 & 0.90          & 22.39 & 0.19 & 0.84 \\
SD-XL  &22.56 & 0.14 & {\bf 0.91}    & 22.60 & 0.17 & 0.84 \\  \hline
\name  &{\bf 23.62}&{\bf 0.09}& {\bf 0.91} &{\bf 23.21}&{\bf 0.11}&{\bf 0.85}\\ 
\hline
\rowcolor{GrayBorder}
        & \multicolumn{6}{c}{HM3D} \\           
RePaint~\cite{lugmayr22repaint} 
       &19.62 &0.24 &0.88           & 18.67& 0.35 & 0.82  \\ 
SD     &22.02 &0.19 &0.89           & 24.32& 0.19 & 0.89 \\
SD-XL  &22.91 &0.14 &0.89           & 24.33& 0.14 & 0.89 \\ \hline
\name  &{\bf 23.34} &{\bf 0.09} &{\bf 0.90}  &{\bf 26.43}&{\bf 0.10}&{\bf 0.90} \\ \hline
\end{tabular}
%\vspace{-0.5mm}
\caption{Quantitative results on MegaDepth, StreetView and HM3D datasets, with both semantic and random masks.}
\vspace{-0.5mm}
\label{tab:3datasets}
%\vspace*{-6mm}
\end{table}
%----------------------------------------------------
\vspace{-1mm}
Tables~\ref{tab:3datasets}-\ref{tab:walking} report quantitative results of comparison to the prior art on HM3D, MegaDepth, StreetView datasets and four scenes from WalkingTour dataset. RePaint~\cite{lugmayr22repaint} clearly underperforms on all datasets, due to mismatch between training and evaluation data. Our method outperforms SD and SD-XL models, trained  for much longer and significantly larger datasets (see Supplementary), with both semantic and random masks. The gain is particularly important in complex real-world scenes, such as WalkingTour dataset.

%-------------------------------------------------------
\begin{table*}[ht!] \centering
\setlength{\tabcolsep}{3pt}
\begin{tabular}{l|ccc|ccc|ccc|ccc}
\specialrule{1pt}{0.5pt}{0.2pt}
\rowcolor{GrayBorder}
Scene  & \multicolumn{3}{|c|}{Amsterdam} & \multicolumn{3}{|c|}{Istabul} & \multicolumn{3}{|c|}{Zurich} & \multicolumn{3}{|c}{Stockholm}\\ \hline
\rowcolor{GrayBorder}
Method&PSNR$\uparrow$&LPIPS$\downarrow$&SSIM$\uparrow$ 
                &PSNR$\uparrow$&LPIPS$\downarrow$&SSIM$\uparrow$
                       &PSNR$\uparrow$&LPIPS$\downarrow$&SSIM $\uparrow$
                            &PSNR$\uparrow$&LPIPS$\downarrow$&SSIM$\uparrow$
\\ \hline
\rowcolor{GrayBorder}
   & \multicolumn{12}{|c}{Semantic Mask} \\
RePaint~\cite{lugmayr22repaint} 
        &10.87&0.56&0.85    &10.17&0.59&0.78    &11.48&0.53&0.89  &12.22&0.51&0.86 \\
SD      &14.09&0.54&0.84    &14.64&0.23&0.77    &15.14&0.21&0.90  &17.41&0.16&0.85 \\
SD-XL   &14.03&0.24&0.84    &13.69&0.24&0.78	&15.17&0.22&0.90  &17.49&0.16&0.86 \\ \hline
\name   &{\bf 19.96}&{\bf 0.09}&{\bf 0.89}    &{\bf 17.30}&{\bf 0.13}&{\bf 0.83}	&{\bf 21.75}&{\bf 0.06}&{\bf 0.93}  &{\bf 19.70}&{\bf 0.09}&{\bf 0.89} \\ \hline
% --------------------------------------
\rowcolor{GrayBorder}
  & \multicolumn{12}{|c}{Random Mask} \\
RePaint~\cite{lugmayr22repaint} 
        &10.68&0.60&0.70   &10.39&0.60&0.70   &11.19&0.58&0.71    &12.37&0.55&0.72 \\
SD      &14.98&0.32&0.70   &16.68&0.26&0.70   &19.87&0.19&0.72    &14.75&0.18&0.73 \\
SD-XL   &14.06&0.24&0.71   &17.75&0.23&0.71   &20.07&0.18&0.73    &14.86&0.18&0.74 \\ \hline
\name   &{\bf 21.36}&{\bf 0.08}&{\bf 0.87}   &{\bf 20.59}&{\bf 0.09}&{\bf 0.85}   &{\bf 23.01}&{\bf 0.06}&{\bf 0.89}    &{\bf 21.42}&{\bf 0.09}&{\bf 0.87} \\
\end{tabular}
\caption{Quantitative results on four scenes from WalkingTour dataset, with semantic and random masks.}
\label{tab:walking}
%\vspace*{-6mm}
\end{table*}

\begin{figure}[t!] \centering   
    \begin{tabular}{cccc}
    Origin & \quad \quad Mask & \quad In-context  & \ Inpainting \\
    \end{tabular}
    \includegraphics[width=0.99\columnwidth]{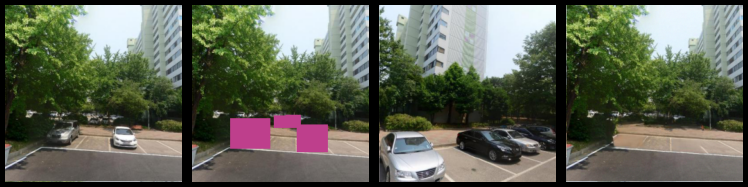}
    \includegraphics[width=0.99\columnwidth]{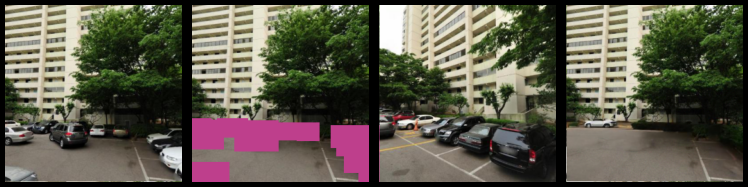}
    \includegraphics[width=0.99\columnwidth]{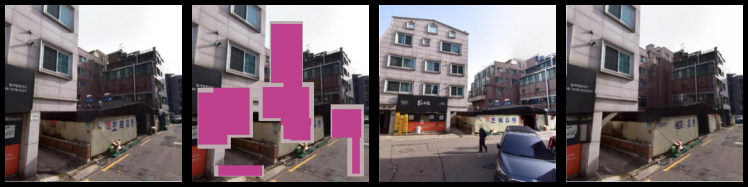}
    \includegraphics[width=0.99\columnwidth]{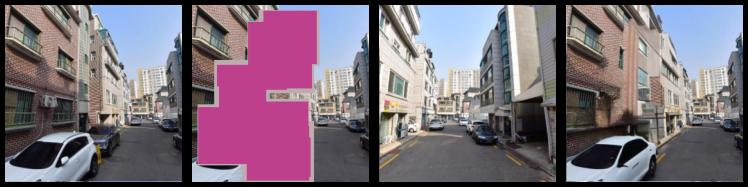}
    \caption{\label{fig:qal_nlk}StreetView dataset: inpainting with semantic (top) and random (top) masks.}    
\end{figure}

%===============================================================
\PAR{Qualitative results}
\label{ssec:qualitative}
We complement the benchmark with visual inpainting examples for all datasets.
In Figures~\ref{fig:qal_nlk}-\label{fig:qal_walk} we present qualitative examples of inpainting for four datasets, with both semantic and random masks. Notice 3D consistency of inpaintings produced by our samples, and the overall harmonized boundaries between the masked ans known regions.

%------------------------------------------

\begin{figure}[t!] \centering    
    \begin{tabular}{cccc}
    Origin & \quad \quad Mask & \quad In-context  & \ Inpainting \\
    \end{tabular}
    \includegraphics[width=0.99\columnwidth]{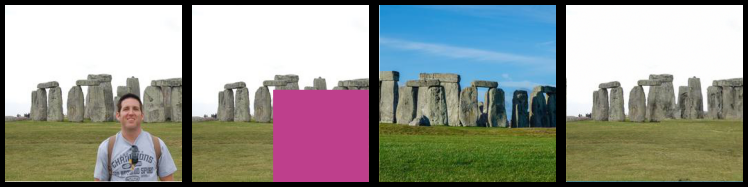}
    \includegraphics[width=0.99\columnwidth]{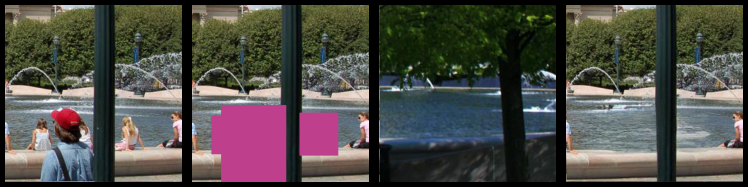}
    \includegraphics[width=0.99\columnwidth]{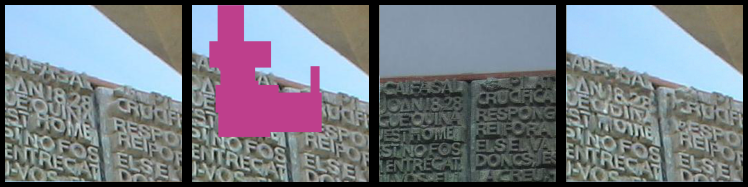}
    \includegraphics[width=0.99\columnwidth]{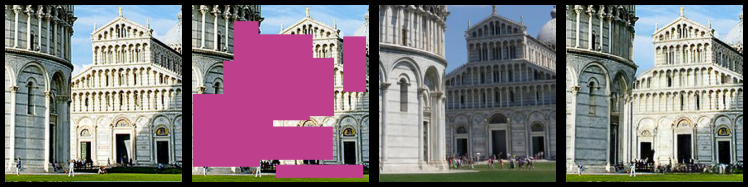}
    \caption{\label{fig:qua_md}MegaDepth dataset: inpainting with semantic (top) and random (bottom) masks.}    
\end{figure}

\begin{figure}[t!] \centering    
    \begin{tabular}{cccc}
    Origin & \quad \quad Mask & \quad In-context  & \ Inpainting \\
    \end{tabular}
    \includegraphics[width=0.99\columnwidth]{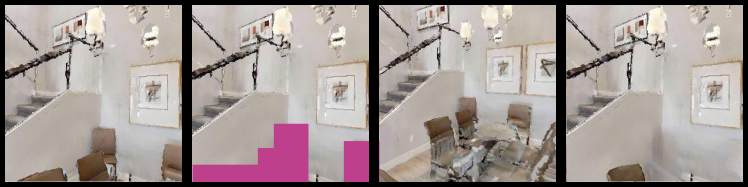}
    \includegraphics[width=0.99\columnwidth]{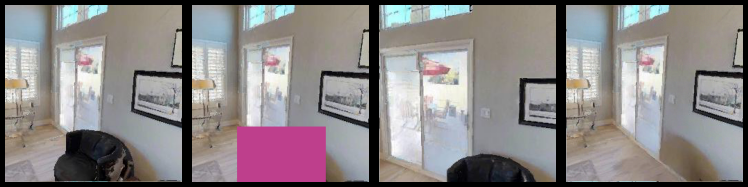}
    \includegraphics[width=0.99\columnwidth]{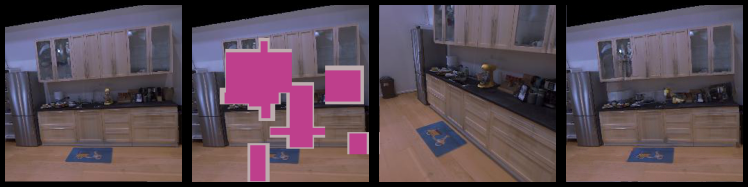}
    \includegraphics[width=0.99\columnwidth]{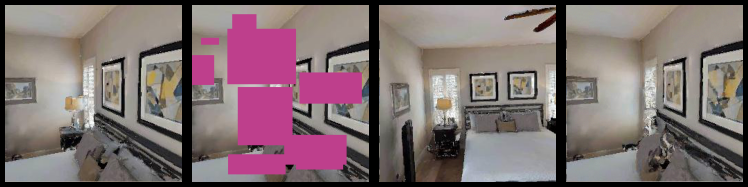}
    \caption{\label{fig:qua_hm3d}HM3D dataset: inpainting with semantic (top) and random (bottom) masks.}
\end{figure}

\begin{figure}[t!] \centering  
    \begin{tabular}{cccc}
    Origin & \quad \quad Mask & \quad In-context  & \ Inpainting \\
    \end{tabular}
    \includegraphics[width=0.99\columnwidth]{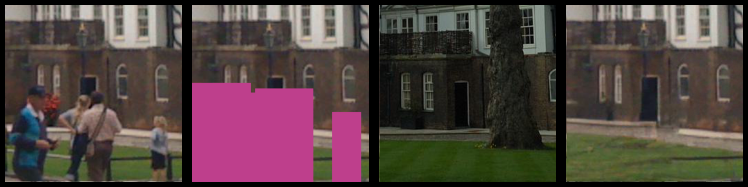}
    \includegraphics[width=0.99\columnwidth]{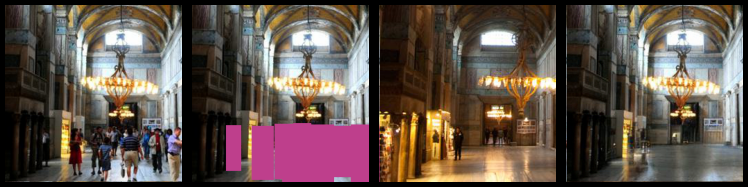}
    \includegraphics[width=0.99\columnwidth]{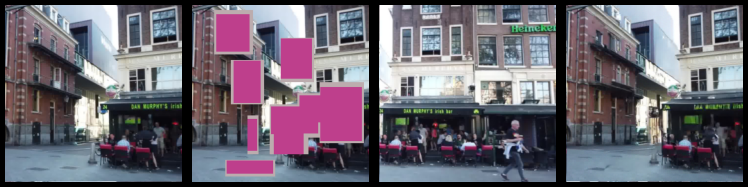}
    \includegraphics[width=0.99\columnwidth]{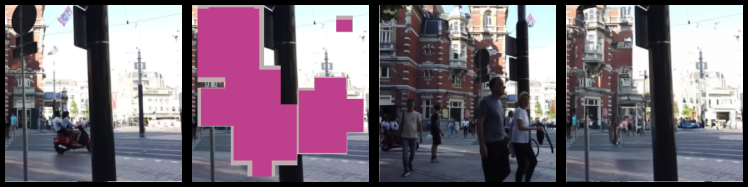}
    \caption{\label{fig:qua_walk} WalkingTour: inpainting with semantic (top) and random (bottom) masks.}    
\end{figure}
%----------------------------------------------
%\vspace*{-0mm}
%================================================================================
\subsection{Ablation studies}
\label{ssec:ablation}

\PAR{Impact of masking ratio}
In evaluation results with random masks in Tables~\ref{tab:3datasets}-\ref{tab:walking}, we generated the mask as a union of $n$=10 random rectangles, with the average mask ratio 0.4. Table~\ref{tab:ratio} below ablates the average mask ratio by running \name \ and SD-XL inpainting on MegaDepth dataset, with additional masking values 0.3, 0.5, 0.6 and 0.7. As the table shows, our \name \ resists better than SD-XL to the larger random mask ratio, and benefits from the in-context images to fill in the masked segments. 

%---------------------------------------------------
\begin{table}[ht!] \centering
\setlength{\tabcolsep}{0.8pt}
\begin{tabular}{l|ccc|ccc}
\specialrule{1pt}{0.5pt}{0.2pt}
\rowcolor{GrayBorder}
Mask  & \multicolumn{3}{c|}{\name} & \multicolumn{3}{c}{SD-XL} \\ 
\rowcolor{GrayBorder}
ratio  & PSNR$\uparrow$ & LPIPS$\downarrow$ & SSIM $\uparrow$ 
                  & PSNR$\uparrow$ & LPIPS$\downarrow$ & SSIM$\uparrow$ \\ \hline
0.3    &{\bf 24.53}&{\bf 0.09} &{\bf 0.89} &24.47 &0.12 &{\bf 0.89}\\ 
0.4    &24.16 & 0.10 & 0.86          & 23.14 & 0.15 & 0.84  \\ 
0.5    &24.02 & 0.11 & 0.84          & 22.08 & 0.18 & 0.80  \\ 
0.6    &23.73 & 0.13 & 0.82          & 21.08 & 0.20 & 0.76  \\ 
0.7    &23.41 & 0.13 & 0.81          & 20.29 & 0.23 & 0.74  \\ 
\hline
\end{tabular}
\caption{Ablation on the average masking ratio for MegaDepth dataset.}
\label{tab:ratio}
\end{table}

%------------------------------------
\PAR{Impact of sampling scheduler}
Evaluation results presented in Tables~\ref{tab:3datasets}-\ref{tab:walking} are reported for the resampling schedule with 1000 steps and 10 jumps, where the resampling from the noisy version of better harmonization between the masked and unmasked regions. Table~\ref{tab:scheduler} shows ablation of both resampling parameters on the HM3D dataset. It reports three main metrics for 250, 500, 1000 steps and 1, 5, 10 jumps. It shows that faster inpainting time comes with a small performance drop, with the number of jumps playing an important role. 

%--------------- schedulers T = 250, 500, 750, 1000...j = 1,5,10, all on HM3D
\begin{table}[h] \centering
\setlength{\tabcolsep}{1.5pt}
\begin{tabular}{l|ccc|ccc|ccc}
\specialrule{1pt}{0.5pt}{0.2pt}
\rowcolor{GrayBorder}
Nsteps          & \multicolumn{3}{c|}{250} & \multicolumn{3}{c|}{500} & \multicolumn{3}{|c}{1000} \\ \hline
\rowcolor{GrayBorder}
Njumps          & 1    & 5    & 10   & 1    & 5   & 10          & 1    & 5  & 10 \\ \hline
SSIM$\uparrow$  & 0.83 & 0.89 & {\bf 0.91} & 0.83 &0.90 & {\bf 0.91} & 0.83 & 0.88 & {\bf 0.91} \\ 
LPIPS$\downarrow$& 0.26 & 0.12 & 0.10 & 0.26 &0.11 &{\bf 0.09} &  0.26 & 0.11  & {\bf 0.09} \\  
PSNR$\uparrow$  &18.7 &23.2 &24.1 &18.7 &23.7&{\bf 25.1}&      18.7 & 23.7 & {\bf 25.1} \\ \hline
\end{tabular}
\caption{Ablation on resampling steps and jumps on HM3D dataset.}
\label{tab:scheduler}
\vspace*{-1mm}
\end{table}

%----------------------------------------------
\subsection{Limitations}
\label{ssec:limitations}
Image inpainting with just an original image often leads to irrealistic off-context and hallucinations generated by diffusion models (some of such examples are shown and compared to \name \ inpaintings in Supplementary). An additional viewpoint of the scene provides in-context guidance and injects 3D priors making inpainting more realistic and consistent. Unfortunately, it can be insufficient for reaching the goal in crowded real-world scenes. As result, inpainting can replace masked obstacles with other obstacles. This requires exploiting multi-view information when parts of the scene are observed in some frames but occluded in others. Extending one additional image by multiple ones will allow to address other challenges, like consistent object removal in videos, without 3D supervision.

%%%%%%%%%%%%%%%%%%%%%%%%%%%%%%%%%%%%%%%%%%%%%%%%%%%%%%%%%%
\section{Conclusion}
\label{sec:conclusion}
\noindent 
We address the problem of 3D inconsistency of image inpainting based on diffusion models. We propose a generative model using image pairs that belong to the same scene. We modify the generative diffusion model by incorporating an alternative point of view of the scene into the denoising process. 
Training unconditional diffusion models with additional images as in-context guidance allows to harmonize masked and non-masked regions while repainting and ensures the 3D consistency. We evaluate our method on one synthetic and three real-world datasets and show that it generates semantically coherent and 3D-consistent inpaintings and outperforms the state-of-art methods.
An additional viewpoint of the scene provides in-context guidance and injects 3D priors into the denoising process, reducing off-context hallucinations and making inpainting more realistic and 3D consistent, without 3D supervision.

{
    \small
    
    \bibliographystyle{ieeenat_fullname}
    \bibliography{bibliography}
}

\end{document}